\documentclass[letterpaper, 10 pt, conference]{ieeeconf} 

\usepackage{url} 
\usepackage{graphicx}
\usepackage{float}
\usepackage{algorithm}
\usepackage{algorithmic}
\usepackage{tablefootnote}
\usepackage{amsmath}
\usepackage{tabu}

\IEEEoverridecommandlockouts      
\title{\LARGE \bf Using  Deep Neural Networks to Translate Multi-lingual Threat Intelligence
}

\author{Priyanka Ranade, Sudip Mittal, Anupam Joshi and Karuna Joshi \\
University of Maryland, Baltimore County, Baltimore, MD 21250, USA\\
Email: $\lbrace$priyankaranade, smittal1, joshi, karuna.joshi$\rbrace$@umbc.edu
}

\begin{document}

\maketitle
\thispagestyle{empty}
\pagestyle{empty}

\begin{abstract}

The multilingual nature of the Internet increases complications in the cybersecurity community's ongoing efforts to strategically mine threat intelligence from OSINT data on the web. OSINT sources such as social media, blogs, and dark web vulnerability markets exist in diverse languages and hinder security analysts, who are unable to draw conclusions from intelligence in languages they don't understand. Although third party translation engines are growing stronger, they are unsuited for private security environments. First, sensitive intelligence is not a permitted input to third party engines due to privacy and confidentiality policies. In addition, third party engines produce generalized translations that tend to lack exclusive cybersecurity terminology. In this paper, we address these issues and describe our system that enables threat intelligence understanding across unfamiliar languages. We create a neural network based system that takes in cybersecurity data in a different language and outputs the respective English translation. 
The English translation can then be understood by an analyst, and can also serve as input to an AI based cyber-defense system that can take mitigative action. As a proof of concept, we have created a pipeline which takes Russian threats and generates its corresponding English, RDF, and vectorized representations. Our network optimizes translations on specifically, cybersecurity data.


\end{abstract}


\section{Introduction}\label{intro}

Information across political, cultural, and geographical boundaries is widely communicated over a global Internet. Today, we have a multilingual Internet where people converse in a variety of languages like English, Mandarin, Russian, Hindi, etc. \cite{worldstats}. Cyber threats in particular, originate from and are mitigated over a broad range of geographic regions. Although a significant amount cybersecurity web data is available, it is spread among major natural languages, decreasing interoperability between multilingual systems. This creates difficulty in employing strong cyber risk management across organizations worldwide. Specifically, amongst state actors or major criminal networks, it is likely that the threat information is in a language other than the language of the analyst. 

Intelligence gathering spans an expansive geographic distribution. As a result, cybersecurity actors, both attackers and defenders, converse over \emph{non-traditional sources} such as social media, blogs, dark web vulnerability markets, etc. in diverse languages. These non-traditional sources are becoming an important asset for threat intelligence mining \cite{attack2017register} and many times are first to receive the latest intelligence about vulnerabilities, exploits, and
threats \cite{vul2017register}. 
The multilingual nature of these non-traditional sources is a potential hindrance for cyber-defense professionals, as they might be limited by their knowledge of different languages. Despite this significant issue, the role of language in addressing cyber threats has been under explored. Multilingual understanding, adds to the many challenges security analysts continue to encounter. The security industry is heavily dependent upon the security analyst's ability in using specialized experience to reason over the disparate pieces of intelligence data available on the web, in order to discover potential threats and attacks. 

The abundance of cybersecurity web data has led to the use of AI/NLP based cyber-defense systems to help analysts extract relevant pieces of information that may constitute an attack. These systems need the ability to process multiple languages to keep up to date with the most current threat intelligence. A multilingual Internet needs a multilingual approach to cybersecurity.

While modern cyber defense systems have the ability to reason over disparate pieces of threat intelligence data on the web, we hope to create a defensive system that also understands various languages, by using the English language as a baseline. In our previous work, we developed \emph{CyberTwitter} and \emph{Cyber-All-Intel} \cite{mittal2016cybertwitter,mittal2017thinking}, systems that mine threat intelligence data from various sources, and automatically issues cybersecurity vulnerability alerts to users. This work extends these cyber-defense systems to a wider spectrum of potential threats, by mining threat intelligence data in a multitude of languages. These systems typically produce ``cyber terminology representations'' \cite{mittal2016cybertwitter,mittal2017thinking} to categorize threat-related words, but only learn representations for English. Consequently, if a certain threat is not gathered under a specific language, the system will not have a representation for it, even if it is a known threat in a different language. We use our multilingual threat intelligence system to align cyber terminology representations of different languages, expanding monitoring capabilities across the globe.

In this work, we create a multilingual translation system that harnesses critical cybersecurity data derived from various natural languages to address the international nature of cyber attacks and assist in defensive cyber operations. Our system optimizes translations particularly for cybersecurity data. Specifically, we investigate semantic representation of multiple languages with a corpus from Twitter, including threats and vulnerabilities in two languages, English and Russian. We  build models to relate the vector space representations in the two languages to translate threat from Russian to English.


\begin{figure}[ht]
\centering
\includegraphics[scale=0.25]{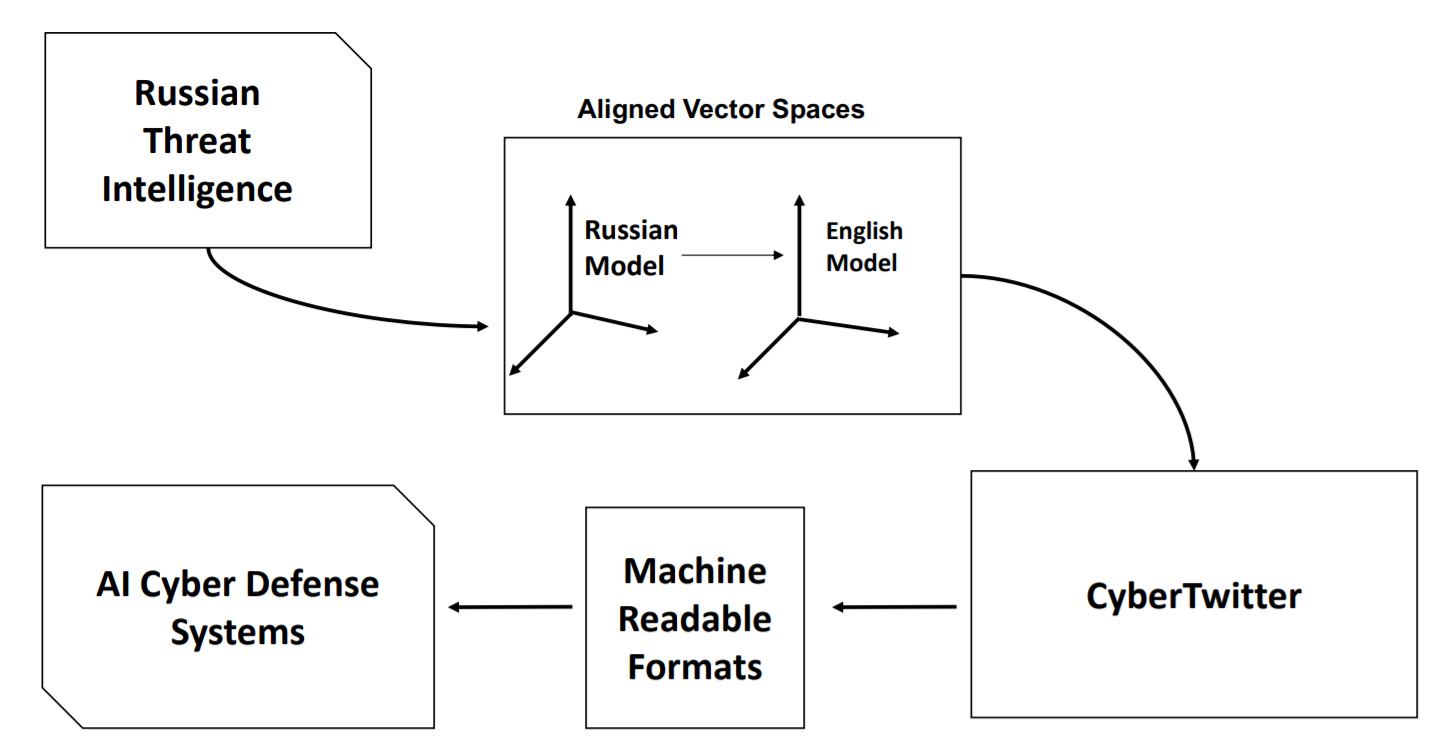}
\caption{Multilingual Threat Intelligence Platform}
\label{fig:arch} 
\end{figure}

Our overall use case (see Figure \ref{fig:arch}), utilizes embeddings created from Russian and English threat intelligence data. The embeddings help us understand security terms in Russian, by aligning semantically similar Russian cyber terms with their English counterparts. The system first begins by gathering relevant Russian threat intelligence data from sources such as Twitter. The data is then assimilated into a vector representation in order to bring semantically similar terms together \cite{mikolov2013efficient}. The data is then fed into CyberTwitter, which converts the English representation of the Russian data into a machine understandable format defined using our UCO Ontology \cite{syed2015uco} in OWL. This helps cyber-defense systems gain intelligence about threats mentioned in the Russian text. The acquired intelligence is then fed into an AI-based cyber defense system that generates conclusions from a cumulation of aggregated threat intelligence data.


An issue with directly converting threats in foreign languages to machine readable formats, is removing the security analyst from the threat inspection process. Providing analysts with raw translations help them reason over and expand upon a new landscape of threats and vulnerabilities. Our system aims to therefore, serve as an augmentation system that helps analysts divert full attention on their primary roles of analyzing and piecing in novel threat information.


The rest of the paper is organized as follows - Section \ref{relwork} contains the related work. We discuss our intelligence translation framework in Section \ref{arch}.  We present our experimental results and evaluations in Section \ref{eval}. We discuss how we use the translation system with an AI based cyber-defense system in Section \ref{usecasepara}. We conclude in Section \ref{conc}.

\section{Related Work}\label{relwork}
In this section, we present related work on the vector space model uses, neural machine translation, AI-based cybersecurity systems, and cybersecurity understanding across different languages.

\subsection{Text Analysis for domain specific tasks}
Text analytics has been utilized in areas such as information retrieval \cite{ray}, 
machine translation \cite{uszkoreit}, and topic detection \cite{cataldi}. These areas are especially useful for domain specific tasks such as cybersecurity. 

\subsubsection*{Vector Space Models}
Vector Space Models, or word embeddings, hasve been used in Natural Language Processing. Words are embedded in a continuous vector space such that, words that appear in the same contexts are semantically related. 
One method that generates embeddings based on word co-occurrence is word2vec \cite{mikolov2013efficient,mikolov2013distributed}. 
Mikolov et al. \cite{mikolov2013distributed},  showed that proportional analogies 
can be solved by finding the vector closest to the hypothetical vector. 
Embeddings have also been utilized in other areas such as word sense disambiguation \cite{globalcontext}, semantic search \cite{semanticsearch}, and discovering inter-linguistic relations in machine translation studies \cite{multiling}. 

Wordnet \cite{wordnet} is a human curated lexical database that groups together synonyms in the English language. Many other versions of Wordnet have been produced, such as ArabicWordNet and ChineseWordnet \cite{worldnetworld}. These lexical databases are often times used to aid lexical and term alignment. 

\subsubsection*{Term Alignment in Vector Spaces}

Analogical Relationships are often times utilized to aid term alignment in vector spaces. Term alignment is known as statistically finding correspondences between words in different groups \cite{sundar}. Plas et al. \cite{Plas} utilize automatic world alignment to find translations from Dutch to one or more target languages. Similarly, Brown et al. \cite{brown} aligned sentences with their translations in two parallel corpora, consisting of French and English. Yang et al. \cite{yang} show how the pattern of the context from word embeddings help to align similar word pairs in other languages. Piantra et al. \cite{pianta} created MultiWordnet, an aligned multilingual database curated to produce an Italian Wordnet, by aligning synonyms in Italian to EuroWordNet. Niemann et al. \cite{niemann} aligned WordNet synonym sets and Wikipedia articles to group article topics based on synonyms.

\subsection{Neural Machine Translation}
Word embeddings have aided in a diversity of machine translation tasks. Neural machine translation typically operates through the encoder-decoder-attention architecture \cite{bahdanau2014}. 
More recently, bilingual word distributions have been trained using unsupervised methods such as Latent Dirichlet allocation (LDA) and Latent semantic analysis (LSA) to aid machine neural translation \cite{bilingual}. 
 Giles et al. \cite{monolingual} trained word embeddings from monolingual data and utilized external and internal vectors as input for the network utilized to train unfamiliar instances of words. In terms of semantic translation tasks, Hill et al. \cite{hill} show that translation-based embeddings work better in applications that require concepts organized according to similarity.


\subsection{Cybersecurity understanding across multiple languages}

Cybersecurity terminology definitions differ across cultures and languages. 
The Department of Homeland Security started developing multilingual resources, to help link cybersecurity understanding across international governments \cite{dhs}. Klavens et al. \cite{klavans} outlines the importance of linguistics in the domain of security and claims language analysis propels understanding of communication between cyber-crime activist groups, filtering relevant data collection, and understanding the intention behind the words.


\subsection{AI-Based Cyber Defense Systems}

The use of social media in threat intelligence mining, provides a new interface between the public and the Intelligence Community. Twitter data in particular, is seen as a reliable OISNT resource due to its real time nature during high impact events, such as terrorist attacks \cite{gupta}. Mittal et al. \cite{mittal2016cybertwitter} developed CyberTwitter, a threat intelligence framework that utilizes twitter data to automatically issue security vulnerability alerts to users.  Similarly, the Cyber-All-Intel system collects OISNT data, stores it in a cybersecurity corpus, and utilizes word vectors for cybersecurity term similarity searches \cite{mittal2017thinking}.

\section{Intelligence Translation Architecture }\label{arch}
In this section, we describe our data collection methods, vector space generation, alignment techniques and neural machine translation framework. We first create a multilingual cybersecurity corpus that contains tweets about threats and vulnerabilities in various languages. In this paper, to create a proof of concept, we focus on English and Russian. Using the collected corpus,we then produce English and Russian vector embeddings. Once we create the embeddings, we align both vector spaces utilizing an alignment database. Once the spaces are aligned, we are able to undertake semantic translation of Russian cyber threats and vulnerabilities to English.

\subsection{Creating a Multi-Lingual Cybersecurity Corpus}\label{corpuspara}



We collect data through the Twitter streaming API using cybersecurity keywords: \textit{XSS, CVE, spam, malware, data, attacker, DNS, DDOS, code, ciphertext, cryptography, hacked, overflow, breach, sniffer, buffer, firewall, domain, hijacking, checksum, virus, vulnerability, arbitrary, protocol,} etc. 
These keywords were suggested by multilingual cybersecurity domain experts \cite{nist} and various security analysts. 
We use the Twitter API language capabilities to detect ``tweet language'' through a flag ($en$=English, $ru$=Russian).  Setting this flag provides us the ability to collect data in both languages. The data is stored and separated by language in MongoDB.

\begin{figure}[h]
\centering
\includegraphics[scale=0.4]{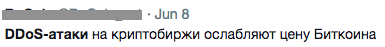}
\caption{Sample tweet from a Russian corporation on crypto DDoS attacks that translates to ``DDos attacks on cryptocurrency weaken the price of Bitcon.''}
\label{fig:tweet} 
\end{figure}

Collecting data using these keywords, gives us a direct interface to Russian cyber colloquialisms. 
For example, the tweet depicted in figure \ref{fig:tweet}, reveals a regional-specific DDoS attack, to threat analysts outside of Russia.


\subsection{Embedding Generation}\label{embeddingpara}
We generate English and Russian word embeddings using Word2Vec \cite{mikolov2013efficient,mikolov2013distributed}. 
We created two separate vector space models from the English tweets and the Russian tweets. 

\begin{figure}[h]
\centering
\includegraphics[scale=0.45]{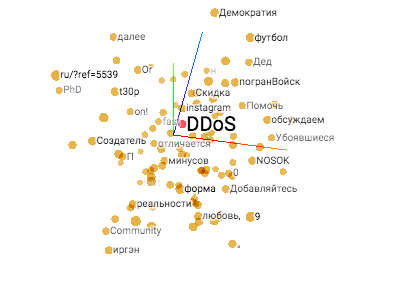}
\caption{Russian Embedding Training Snapshot around the word ``DDOS''. Neighboring words include community and proxy.}
\label{fig:ddosembedding} 
\end{figure}

In our system, these models are used for semantic translation of Russian cybersecurity text to English (see Section \ref{aligndb}). Words in the embedding space are \emph{semantically similar} if grouped together around the same neighborhood. For example, in our English model, words like, malware, virus, trojan, etc. will be clustered together. Figure \ref{fig:ddosembedding}, depicts a 20th iteration training snapshot, of Russian words that start appearing near ``DDoS'', a type of cybersecurity attack.

\subsection{Intelligence Translation Framework}\label{translation}

In this section, we describe our intelligence translation framework that creates many-many mappings among cybersecurity terms in Russian and English. We use the word embeddings produced in Section \ref{embeddingpara}, and create an alignment database, later used as the dataset for training our neural network described in Section \ref{rnnpara}.

In a high level usecase, a cyber-defense system like CyberTwitter \cite{mittal2016cybertwitter}, will take as its input Russian threat intelligence and create machine readable threat intelligence. This scenario is discussed in detail in Section \ref{usecasepara}.
\subsubsection{Creating an Alignment Database} \label{aligndb}
\begin{figure}
\centering
\includegraphics[scale=0.27]{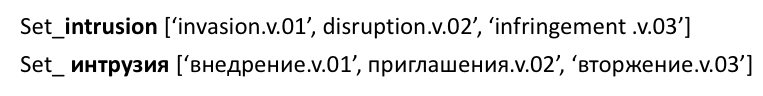}
\caption{English and Russian Synset Examples for ``intrusion''.}
\label{fig:synset} 
\end{figure}

In order to create relationships between English and Russian cybersecurity words, we created a dataset to align the English and Russian vector embeddings. 
An alignment in our system means creating true positive mappings of Russian cybersecurity terms to their English counterparts. 
We derived \emph{cybersecurity synsets} for the Russian and English vocabulary embeddings, created in section \ref{embeddingpara}.
These cybersecurity synsets include contextually similar words to each vocabulary word in the Russian and English vector spaces. We emphasize that, when we say contextually similar words, we bring together cybersecurity terms in the same word sense.  
The lexical database Wordnet \cite{wordnet}, groups similar words into sets of synonyms called ``synsets''. WordNet does not support the Russian language. We found a similar lexical database called Russnet \cite{russnet}, specifically for the Russian language. We utilize the English synsets provided by WordNet, and the Russian synsets provided by Russnet to create our cybersecurity synsets. An example of a cyber synset we derived is shown in Figure \ref{fig:synset}.

We tasked three native Russian speakers, who served as annotators, to manually verify the quality of cybersecurity synsets produced. We use the Cohen's kappa to compute the inter-annotator agreement, and keep only those cybersecurity synsets that scored higher than 0.66. The annotators confirmed that the synsets in Wordnet, and the synsets in Russnet, were not only similar on a translation level, but also semantically similar in a cultural context. 


We use these cybersecurity synsets to then create a neural network, that given an input Russian embeddings, outputs its English equivalent. 

\subsubsection{Intelligence Translation Network}\label{rnnpara}


\begin{figure}
\centering
\includegraphics[scale=0.27]{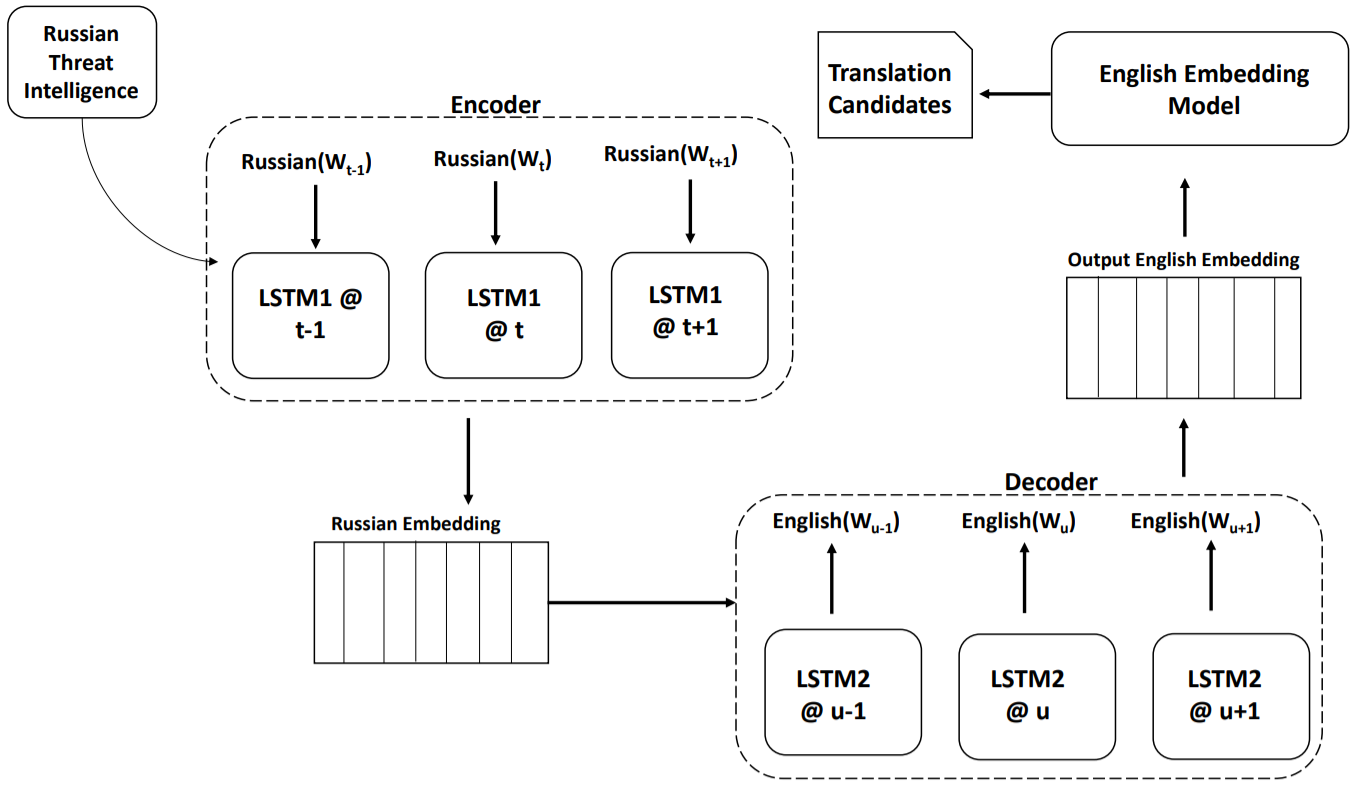}
\caption{Intelligence Translation Architecture}
\label{fig:rnnarch} 
    \vspace{-6mm}

\end{figure}

Our intelligence translation architecture is shown in Figure \ref{fig:rnnarch}. We implement a 
encoder-decoder network, which is a dual Recurrent Neural Network (RNN). The encoder serves as an input RNN and the decoder, serves as the output.  
The encoder-decoder architecture projects the input Russian word to be translated into the English embedding space, by returning words with a representation closest in the English vocabulary.

The encoder-decoder network is implemented using a ``sequence-to-sequence architecture'' \cite{neuralmachineone,neuralmachinetwo}. The encoder-decoder network is able to process past and future words in a sequence, and is also able to map an input sequence to an output sequence of a different length. 
					
The model uses word embeddings produced in section \ref{embeddingpara}. We initialize Russian embeddings as the input in the encoder state, and the English embeddings as the output of the decoder state. We utilize the cybersecurity synsets, from section \ref{aligndb}, as the training set. Our hyperparameters were set as, batch size = 64, epochs = 100, latent dimentionality = 256, sample number = 10,000.


					
The encoder and decoder utilize a Long Short Term Memory (LSTM) cell \cite{neuralmachineone}. In the hidden layer, we have one dense layer, with a softmax activation function, which allows the model to learn a mapping from the Russian vector representation, to the English vector representation. 
The encoder, takes in Russian words and maps them to their respective vector representations. The decoder then, creates a translation of the input word and generates its predicted aligned semantic English embedding. We discuss the accuracy for this model in Section \ref{eval}.


\section{Evaluation}\label{eval}
In this section, we describe our experimental setup and evaluate our intelligence translation system. We first measure our translation precision through BLEU and accuracy metrics. Later, we compare our system against other commercial translation services. 

\subsection{Accuracy and BLEU score}

We first evaluate our encoder-decoder architecture through an accuracy metric and a BLEU (Bilingual Evaluation Understudy) score (see Table \ref{table:eval}). The accuracy metric computes the percentage of times that predictions match labels. BLEU scores, are standard metrics for evaluating a generated translation to a reference word \cite{neuralmachinetwo}. An accuracy above ``60\%'', perplexity under ``6'', and a BLEU Score between ``15 and 36'' is considered robust \cite{neuralmachinetwo}.

\begin{table}[ht]
	\centering
    \begin{tabular}{|l|l|}
    \hline
    \textbf{Measure}   & \textbf{Value}   \\ \hline
    Accuracy   & 97.22\% \\ \hline
    Perplexity & 4.07    \\ \hline
    BLEU Score & 28.4    \\ \hline
    \end{tabular}
    \label{table:eval}
    \caption{Evaluation Metrics}
    \vspace{-10mm}
\end{table}


\subsection{Measuring against Commercial Systems}

We measure the precision of our translations by checking a randomly generated sample of the output against Google Translate\footnote{\url{https://translate.google.com}}. We proved that our system produces more effective translations for the security domain. 
We extracted 1000 randomly selected tweet translations and compared the output against the Google Translate API. 
We check our translations against the ones provided by Google Translate, both syntactically and semantically. 
On evaluating 1000 random samples, there was a 64.3\% syntactic correlation between the two systems. 

\begin{figure}
\centering
\includegraphics[scale=0.25]{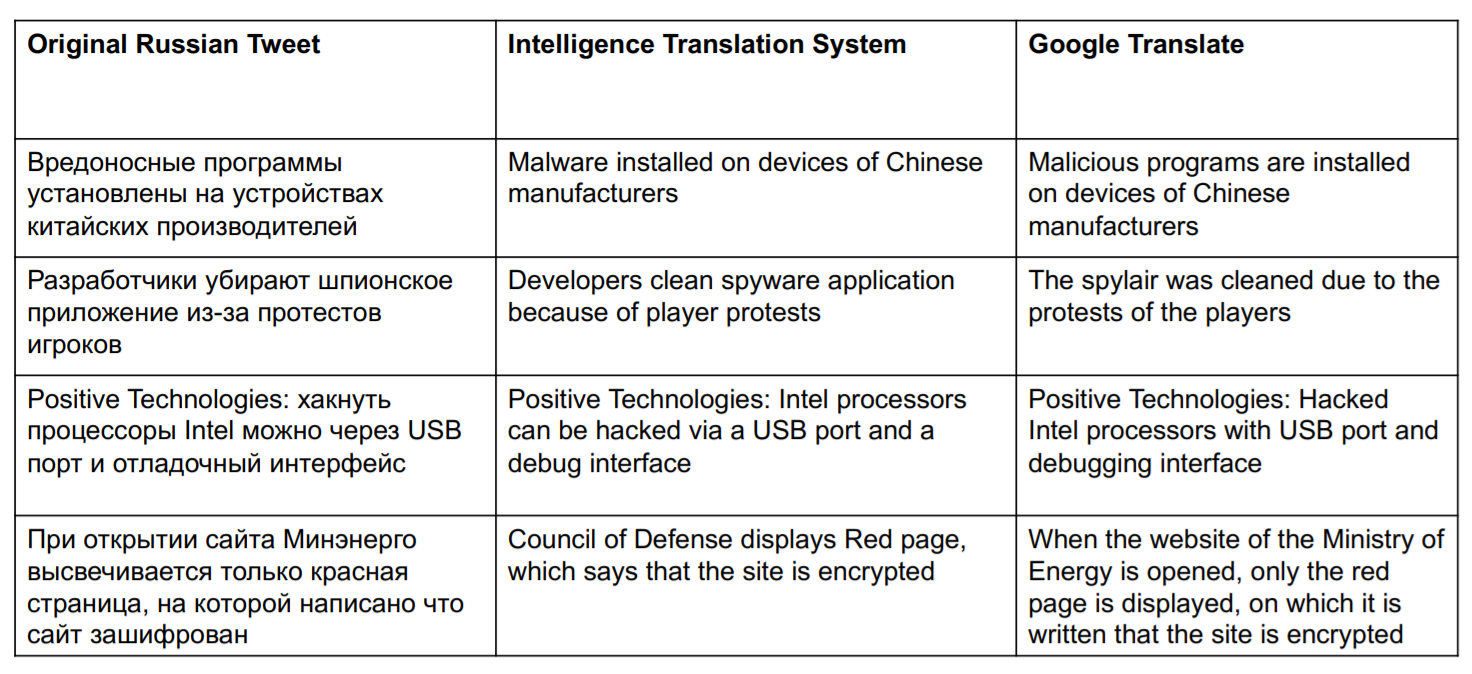}
\caption{Tweet Translation Samples.}
\label{fig:resulttweet} 
\vspace{-6mm}
\end{figure}

We further evaluated the 357 samples that were not syntactically equivalent and tasked two security analysts to manually evaluate the semantic meanings of the translation outputs. We found that of the 357 outputs, 349 were semantically similar, but not syntactically similar to the Google translation, showing 97\% semantic relevance. The annotators concluded our translations are preferable through a security perspective, in that they proliferate terms unique to the security industry. The commercial translation services are generalized while our system is domain specific. 
These security specific translations can be attributed to the architecture of our model, that utilizes a specialized aligned database made with relevant cybersecurity mappings. Examples of unequal but semantically similar translations in our system and Google Translate are listed in figure \ref{fig:resulttweet}. In example 1, ``malware'' registers more with a security analyst than ``malicious programs''. In example 2, the Google Translate system translated the relevant Russian text as ``spylair'', while our system gives the correct translation as ``spyware''. These are clear instances in which our translation will provide more relevant and direct intelligence for a security professional.

Another benefit that our system provides is that it can run independently in secluded operational settings. A security analyst may not be able to input their sensitive data into third party platforms due to privacy, security, and confidentiality policies. 

\section{Use by Cyber-Defense Systems}\label{usecasepara}

Web based unstructured, textual sources such as Twitter, Reddit, blogs, dark web forums, etc. provide a rich multilingual source of information about cyber threats and attacks. In addition to providing details of existing attacks, such sources (especially the dark web) can serve as advance indicators of attacks in terms of discussions around newly discovered vulnerabilities. This information is available in textual sources traditionally associated with Open Sources Intelligence (OSINT), as well as in data that is present in hidden sources like dark web vulnerability markets. 

\begin{figure}
\centering
\includegraphics[scale=0.36]{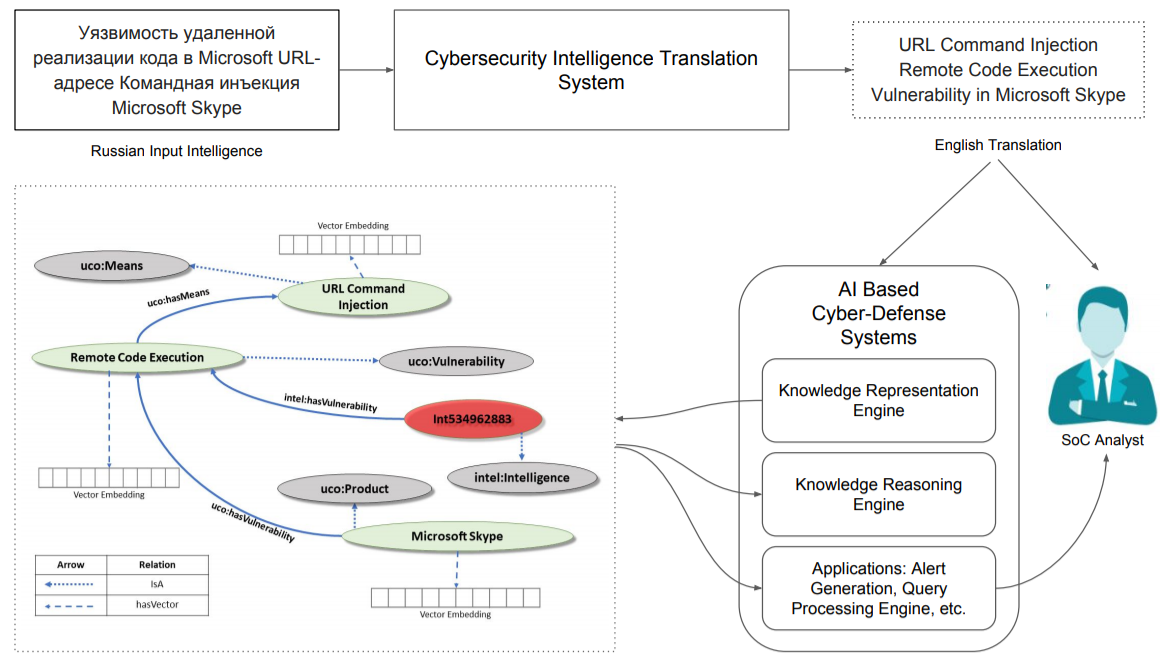}
\caption{Using the intelligence translation system with an AI based cyber defense system.}
\label{fig:aiusecase} 

\end{figure}

The intelligence translation system that we discuss in Section \ref{arch} will help us automate this process by taking data from a variety of multilingual sources, extracting, representing and integrating the knowledge present in it as embeddings and knowledge graphs, and then use the resulting artificial intelligence systems to provide actionable insights to SoC professionals. Figure \ref{fig:aiusecase} showcases our pipeline, which takes in Russian threat intelligence and stores it in as a \emph{VKG structure} \cite{mittal2017thinking}.

Two such systems that we have developed in the past are CyberTwitter \cite{mittal2016cybertwitter} and Cyber-All-Intel \cite{mittal2017thinking}. The systems store threat intelligence in a knowledge representation that can be used by AI based cyber-defense systems (See Figure \ref{fig:aiusecase}). Such systems generally have a knowledge representation engine, a reasoning engine, and few applications like an alert generation system, recommender system, query processing system, etc. 

The knowledge representation system, converts input threat intelligence (usually in a textual format) into a machine readable format. In our system we represent it in RDF\footnote{\url{https://www.w3.org/RDF/}}, with cybersecurity domain knowledge provided by the Unified Cybersecurity Ontology (UCO) \cite{syed2015uco}. The intelligence ontology \cite{mittal2016cybertwitter} provides information about the intelligence domain. We also include specific conceptual embeddings for security concepts in our threat representation format \cite{mittal2017thinking}. The knowledge reasoning part of the system provides domain specific reasoning capability generally encoded as logical rules by a domain expert. The applications and the reasoning engine generally use the machine readable representation to provide specific functionality. Figure \ref{fig:aiusecase} also provides the graph structure for the translated English intelligence: ``\emph{URL Command Injection Remote Code Execution Vulnerability in Microsoft Skype}''. Figure \ref{fig:exampleRDF} provides the RDF representation for the same intelligence.

\begin{figure}[!htbp]
\small
\begin{minipage}{1\columnwidth} 

@prefix uco: $<$http://accl.umbc.edu/ns/ontology/uco\#$>$ . \\
@prefix intel: $<$http://accl.umbc.edu/ns/ontology/intelligence\#$>$ . \\
@prefix rdf: $<$http://www.w3.org/1999/02/22-rdf-syntax-ns\#$>$ . \\
@prefix rdfs: $<$http://www.w3.org/2000/01/rdf-schema\#$>$ . \\
@prefix xml: $<$http://www.w3.org/XML/1998/namespace$>$ . \\
@prefix xsd: $<$http://www.w3.org/2001/XMLSchema\#$>$ . \\
@prefix dbp: $<$http://dbpedia.org/resource\#$>$ . \\
@prefix owl: $<$http://www.w3.org/2002/07/owl\#$>$ . \\
\\
$<$Int534962883$>$ a intel:Intelligence ; \\
    intel:hasVulnerability $<$remote\_code\_execution$>$ ;\\
\\
$<$command\_injection$>$ a uco:Means .\\
\\
$<$Microsoft\_Skype$>$ a uco:Product ;\\
	uco:hasVulnerability $<$remote\_code\_execution$>$ ;\\
    owl:sameAs dbp:Skype .\\
\\
$<$remote\_code\_execution$>$ a uco:Vulnerability ;\\
    uco:affectsProduct $<$Microsoft\_Skype$>$ ;\\
    uco:hasMeans $<$command\_injection$>$ ;\\
    owl:sameAs dbp:remote\_code\_execution .\\
\end{minipage}
\caption[RDF example.]{RDF for textual input ``URL Command Injection Remote Code Execution Vulnerability in Microsoft Skype''. Also, $owl:sameAs$ property has been used to augment the data using an external source `DBpedia' \cite{auer2007dbpedia}.}
\label{fig:exampleRDF}
\end{figure}

\section{Conclusion \& Future Work}\label{conc}
In this paper, we described the design, implementation, and evaluation of a multilingual threat intelligence translation system. The system uses Russian and English word embeddings created from cybersecurity data, an aligned cyber term database, and a LSTM based neural machine translation architecture, to translate cybersecurity text from Russian to English. With the help of Russian speaking cyber analysts, we created an alignment database by generating synonyms for the Russian and English corpus vocabularies, along with their respective translated Russian and English words. We utilize this database in neural machine translation, where we use an encoder-decoder architecture to map unfamiliar Russian cyber inputs to their English counterparts. We show that our model not only has high syntactic correlation to third party translation systems, but also registers prevalent cybersecurity terms in translation better than third party engines. We extend third party translation systems by creating a domain specific model that can provide more pertinent intelligence for an analyst. Our system can be utilized in private operational settings that do not permit the use of third party applications when dealing with sensitive intelligence data.


A weakness of our system, is the requirement of a cybersecurity rich alignment to train the model. Although we derived a Russian and English cybersecurity synonym sets in this proof of concept, it is an expensive task that will take dispersed effort across the linguistic and security communities, to derive across many other languages.  
In order to create more mappings for cyber terms across other languages like, Mandarin, Cantonese, Portuguese, Arabic, Hindi, etc. future research can include creation of multilingual cyber alignment databases. We can also consider transferring knowledge from languages with an abundance of intelligence to other unknown languages with no or few alignments. We expect that aligned cyber embeddings across many languages can promote international incident response collaboration.
\section*{Acknowledgement}\label{ack}
The work was partially supported by a gift from IBM Research, USA and the Undergraduate Research Award from the University of Maryland, Baltimore County.

\bibliographystyle{plain}
\bibliography{priyanka}
\end{document}